\documentclass[10pt, conference]{IEEEtran}
\IEEEoverridecommandlockouts

\usepackage{cite}
\usepackage{amsmath,amssymb,amsfonts}
\usepackage{algorithmic}
\usepackage{graphicx}
\usepackage{textcomp}
\usepackage{xcolor}
\usepackage{soul}
\usepackage{svg}
\usepackage{subcaption}
\def\BibTeX{{\rm B\kern-.05em{\sc i\kern-.025em b}\kern-.08em
    T\kern-.1667em\lower.7ex\hbox{E}\kern-.125emX}}

\usepackage[
placement=top,
angle=0,
color=red,
scale=1.1,
hshift=0,
vshift=-15
]{background}
\backgroundsetup{contents=\bf{----------------------------------------------------------------- PREPRINT -----------------------------------------------------------------}}
\begin{document}
% my commands
\def\NoNumber#1{{\def\alglinenumber##1{}\State #1}\addtocounter{ALG@line}{-1}}
\definecolor{mygreen}{RGB}{0,204,102}
\newcommand{\ACe}[1]{\textcolor{red}{#1}}
\newcommand{\rev}[1]{\textcolor{blue}{#1}}
\title{\vspace*{18pt}Integrated Hardware and Software Architecture for Industrial AGV with Manual Override Capability}
\author{
    \IEEEauthorblockN{Pietro Iob\IEEEauthorrefmark{1}\IEEEauthorrefmark{2}, Mauro Schiavo\IEEEauthorrefmark{2}, Angelo Cenedese\IEEEauthorrefmark{1}}
    \IEEEauthorblockA{\IEEEauthorrefmark{1}\textit{Department of Information Engineering, University of Padova, Padua, Italy}
    \\\ pietro.iob@phd.unipd.it, angelo.cenedese@unipd.it}
    \IEEEauthorblockA{\IEEEauthorrefmark{2} \textit{Techmo Car S.p.a., Padua, Italy}
    \\\ pietro.iob@techmo.it, mauro.schiavo@techmo.it}
}
% make the title area
\maketitle
\begin{abstract}
This paper presents a study on transforming a traditional human-operated vehicle into a fully autonomous device. By leveraging previous research and state-of-the-art technologies, the study addresses autonomy, safety, and operational efficiency in industrial environments. Motivated by the demand for automation in hazardous and complex industries, the autonomous system integrates sensors, actuators, advanced control algorithms, and communication systems to enhance safety, streamline processes, and improve productivity. The paper covers system requirements, hardware architecture, software framework and preliminary results. This research offers insights into designing and implementing autonomous capabilities in human-operated vehicles, with implications for improving safety and efficiency in various industrial sectors.
\end{abstract}
\IEEEpeerreviewmaketitle
\section{Introduction}
Recent advancements in autonomous systems have revolutionized various industries, enhancing efficiency, safety, and productivity. This paper focuses on developing autonomous vehicles for industrial settings, specifically transforming a traditional human-operated vehicle into a fully autonomous system \cite{sell2019integration}. The study includes both software and hardware architectures, applied to an Autonomous Ground Vehicle (AGV) prototype for the primary aluminum industry.%, shown (\ref{fig:Render}).

% \begin{figure}[ht]
% \centering{\includegraphics [width=0.48\textwidth] {Figures/Feeding.jpeg}}
% \caption{Rendered image of the vehicle of interest. Courtesy of Techmo Car S.p.a.}
% \label{fig:Render}
% \end{figure}

The motivation behind this research is the increasing demand for automation in industries with hazardous conditions and complex tasks. The primary aluminum industry, reliant on heavy machinery and high-risk operations, serves as a compelling case study. Introducing autonomous capabilities aims to enhance safety, streamline processes, and improve productivity.

The transformation of human-operated vehicles into autonomous devices involves developing a general-purpose platform, considering both software and hardware architectures. While software aspects are well-studied ( \cite{broy2006challenges}, \cite{staron2021automotive}, \cite{serban2018standard}, \cite{berntorp2018control}), the hardware side requires more attention, since, to the knowledge of the author, previous studies focused on academic \cite{fernandes2022design} or specific applications \cite{etezadi2024comprehensive}, lacking a more general-purpose solution which this paper aims to address.

The paper is structured as follows: Section \ref{sec:Requirements} outlines system requirements; Section \ref{sec:SWARch} discusses the software framework; Section \ref{sec:HWARch} details the hardware architecture; Section \ref{sec:FSM} focuses on strategies for integrating manual and automated control; Section \ref{sec:Results} presents experimental results; and Section \ref{sec:Conclusion} summarizes key findings and future research directions.

%In summary, this paper contributes to industrial autonomous systems by detailing the design and implementation of autonomous capabilities in the primary aluminum industry, aiming to drive innovation, enhance safety, and improve operational efficiency.
%
\section{System Requirements}
\label{sec:Requirements}
Autonomous systems necessitate a diverse array of sensors and actuators to operate effectively, a concept well-established in existing literature \cite{ignatious2022overview}, \cite{ayala2021sensors}. This study endeavors to convert a conventional human-operated vehicle into a fully autonomous apparatus capable of executing assigned tasks. Notably, the uniqueness of this project lies in its requirement for manual override capability. Consequently, the system must be outfitted with actuators controllable both by an operator and by a processing unit, configured in a drive-by-wire arrangement.

This section serves to delineate the primary requirements essential for the system's functionality.
\begin{enumerate}
    
    \item Human operator control: Traditionally, vehicles in the primary aluminum industry use hydraulic systems powered by diesel engines due to time-varying electromagnetic fields in their environment. However, due to increasing interest in electric vehicles (EVs), the hydraulic system in this prototype is driven by an electric motor. The prototype includes:
    \begin{itemize}
        \item An electric traction motor with an inverter, controllable via Control Area Network Bus (CAN-BUS) or throttle pedal.
        \item An electro-hydraulic steering valve with a compatible controller, offering control via CAN-BUS or manual actuation.
        \item An electro-hydraulic valve for arm joints, managed via CAN-BUS only.
    \end{itemize}
    \item Drive-by-Wire Capabilities: This requirement highlights the system's ability to be electronically controlled, exclusively through CAN-BUS
    \item Mode Switching Capabilities: The system necessitates a supervisory function to govern the behavior of actuators and their respective controllers. This requirement guarantees the coexistence of both human-operated and automated modes within the same vehicle.
    \item Perception and Localization Capabilities: An autonomous system requires a suite of sensors to perceive its environment for safety (obstacle detection), localization, and mapping. Operating within a known plant environment, the vehicle benefits from environmental familiarity. However, a mapping solution is incorporated to dynamically update the environment and optimize route navigation to designated destinations.
\end{enumerate}
Building upon these fundamental requirements, the subsequent section will delve into the intricate hardware architecture designed to realize these functionalities.
\section{Software Architecture}
\label{sec:SWARch}
%
%The comprehensive exploration undertaken in this study delves deeply into the intricate analysis and fusion of both hardware and software components within a unified system architecture paradigm. This convergence is pivotal in realizing a holistic system capable of seamlessly responding to operator commands. This paper seeks to elucidate the essential procedural steps necessary for integrating these architectural elements, culminating in the development of a fully functional system.

\begin{figure}[ht]
\centering
\includegraphics[width=.43\textwidth]{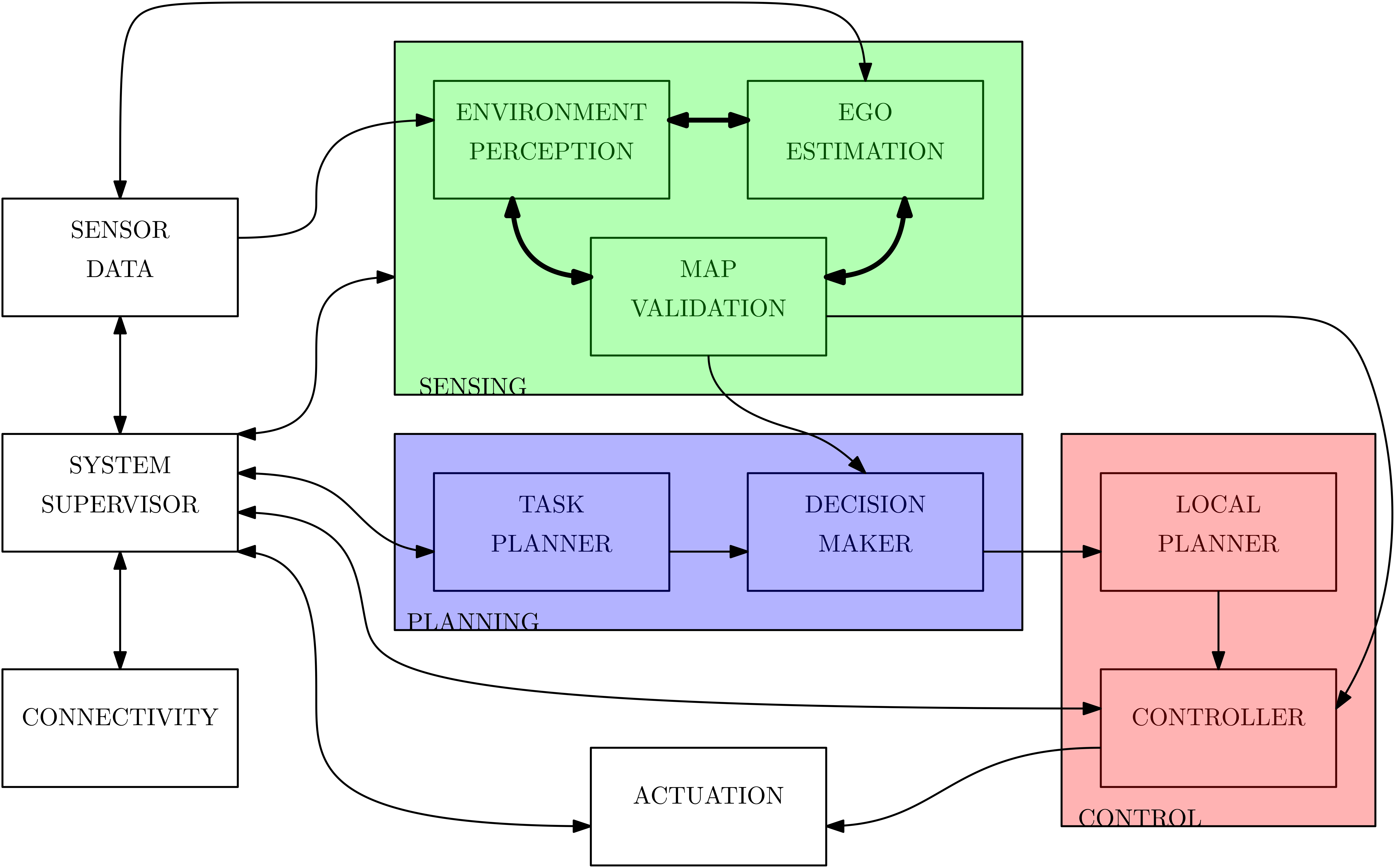}
\caption{Software Architecture of the system, illustrating relevant interconnections.}
\label{fig:SWArchitecture}
\end{figure}

Figure (\ref{fig:SWArchitecture}) provides a schematic representation of the software architecture implemented for the Techmo Car prototype.

The software architecture, as depicted in the aforementioned block scheme, can be segmented into three primary blocks: Sensing, Planning, and Control. These blocks serve as the intermediary between software and hardware components, encompassing sensors, actuators, connectivity, and notably, the system supervisor, which will be further elaborated upon in the subsequent section.

While the Sensing, Planning, and Control components have been extensively addressed in the literature, this section provides a concise overview without delving into specific details. 

\begin{itemize}
\item \textbf{Sensing} \cite{bresson2017simultaneous}, \cite{kumar2023survey}, \cite{yu2020study}: This module processes incoming sensor data and relays relevant information to the planning and control subsystems. It includes three subcomponents: Environment Perception, Map Validation, and Ego Estimation. Environment Perception detects obstacles, markers, pedestrians, and environmental changes. Ego Estimation handles system localization and state determination, including vehicle speed, acceleration, steering angle, joint positions, and system orientation. Map Validation maps the environment, providing essential spatial memory.

\item \textbf{Planning} \cite{gonzalez2015review}, \cite{costa2019survey}: This module manages tasks from the system supervisor and shapes system behavior based on data from the sensing module. It has two subsystems: the task planner, which designs global system behavior, and the decision maker, which relays this to the control system and intervenes during environmental changes or emergencies.

\item \textbf{Control} \cite{veres2011autonomous}, \cite{parekh2022review}: The control module amalgamates information from the sensing and planning subsystems, generating inputs for the actuators. Fundamentally, this module ensures alignment between the designated system task, system dynamics, and the vehicle's surroundings.
\end{itemize}

The proposed architecture embodies flexibility, enabling users to select and implement various algorithms without necessitating system modifications, thus imbuing the architecture with a level of modularity.

\section{Hardware Architecture}
\label{sec:HWARch}
%
%Differently from the Software Architecture presented in the previous section, its Hardware counterpart will be focused on the case study presented in the introduction. However this choice, does not interfere with the generality of this work, since there is the possibility to substitute each and every components, from the sensor to the actuators, maintaining the same high level structure presented in this study.

At the core of this integration process is the delineation of system interconnections, involving sensors, actuators, and controllers. The initial step is defining the network of dependencies among these components. System controllers are divided into two entities: the Autonomous Driving Processing Unit (ADPU) and the Programmable Logic Controller (PLC). The ADPU handles localization, navigation, and overall system control, while the PLC serves as a low-level controller and supervisor.

\begin{figure}[ht]
\centering{\includegraphics [width=.40\textwidth] {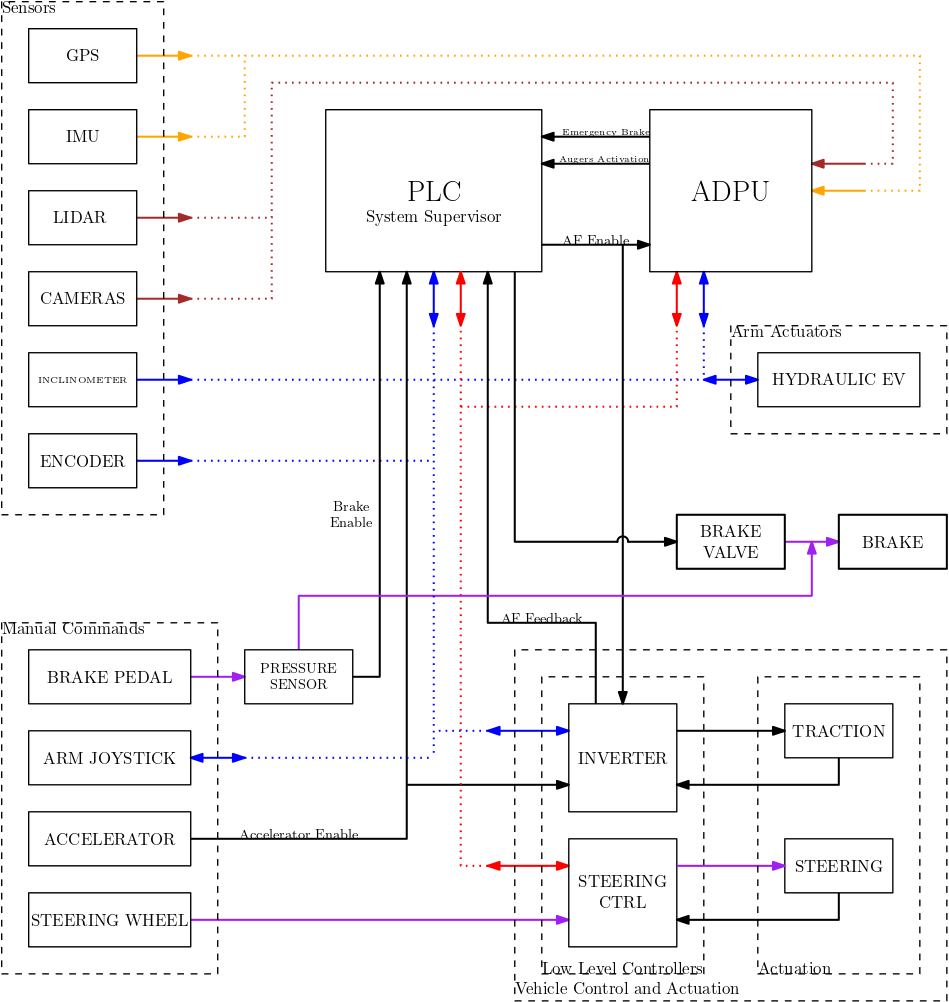}}
\caption{Hardware Architecture of the system, with the relevant interconnections.}
\label{fig:HWArchitecture}
\end{figure}

\begin{figure}[ht]
\centering{\includegraphics [width=.15\textwidth] {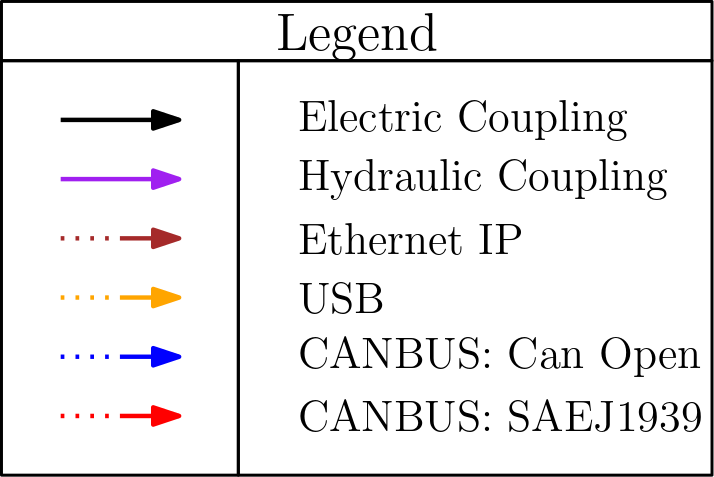}}
\caption{Legend of the interconnection of the system components.}
\label{fig:HWArchitecture_Legend}
\end{figure}

Figure (\ref{fig:HWArchitecture}) provides an overview of the hardware architecture, detailing the interconnections, explained in the legend in Figure (\ref{fig:HWArchitecture_Legend}). %The architecture includes various interconnection types, such as analog signal conduits for direct control of actuators and sensor data relay, represented by continuous lines. The system's communication relies on CAN-BUS, essential for its operational needs.

Directly connecting specific sensors to the ADPU marks a significant shift from traditional methods, driven by a detailed understanding of the system's architecture. This method avoids integration with the PLC, as the data from these sensors is only relevant to ADPU operations, simplifying system interconnectivity, allowing for greater autonomy in sensor selection based on their capabilities rather than PLC compatibility and enhancing system efficiency and robustness.

Figure \ref{fig:HWArchitecture} illustrates the system's core components, featuring sensors like GPS and IMU for precise localization, and Lidars and Cameras for environment mapping and obstacle detection, with Lidars also aiding in localization \cite{graeter2018limo}. Encoders and inclinometers accurately measure arm joint positions.

The system includes various actuators, such as traction and steering systems, each with low-level controllers, and solenoid valves for robotic arm motion control. Manual commands can directly control actuators, allowing immediate operator intervention.

Both the system supervisor and processing unit can control all actuators, with the PLC serving as a safety mechanism. This redundancy enhances safety and system adaptability. Manual commands can be replaced with safety sensors, enabling the System Supervisor to autonomously detect and act on critical situations without changing the system architecture.
\section{Finite State Machine}
\label{sec:FSM}

\begin{figure}[ht]
\centering{\includegraphics [width=.3\textwidth] {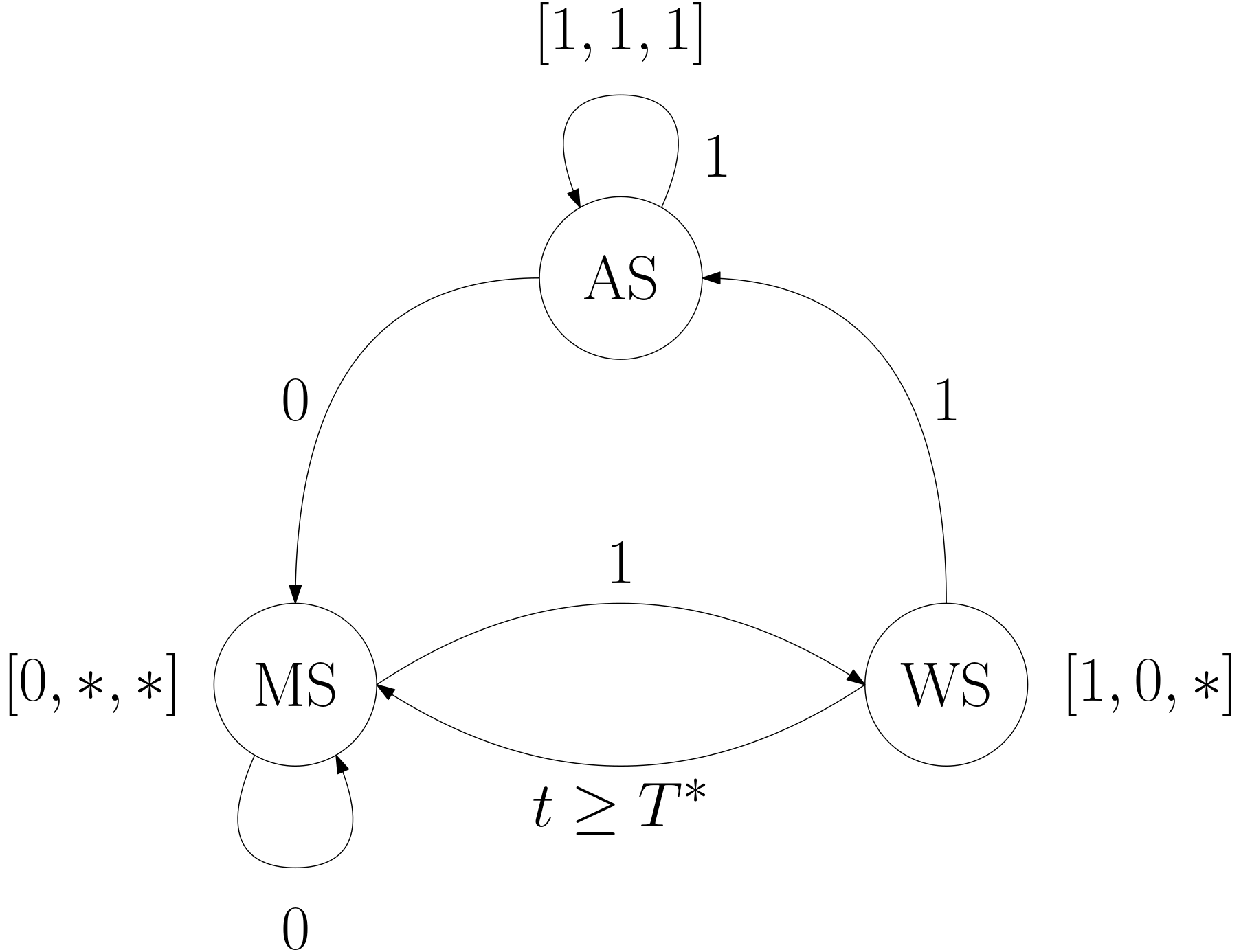}}
\caption{Finite State Machine of the System Supervisor.}
\label{fig:HWDiscreteStateMAchine}
\end{figure}

This section outlines the safety software facilitating interaction between operator control and autonomous driving within the vehicle. A simplified depiction of the Finite State Machine (FSM) in the prototype vehicle is shown in Figure (\ref{fig:HWDiscreteStateMAchine}).

The FSM consists of three primary states:

\begin{itemize}
\item Manual State (MS): Initially, the operator has full control, and communication from the Autonomous Driving Processing Unit (ADPU) to actuators ceases.
\item Waiting State (WS): An intermediary phase evaluating ADPU readiness to assume control.
\item Autonomous State (AS): Signifies autonomous driving mode, with actuators responding to ADPU directives.
\end{itemize}

The FSM defaults to MS at vehicle ignition, permitting consent-based state transitions. User consent operates on a binary principle: $1$ allows progression to AS, $0$ reverts to MS. Upon consent, the system transitions to WS, requesting ADPU control. If confirmed within the timeframe, the vehicle shifts to AS; otherwise, it returns to MS automatically.

\begin{figure}[ht]
\centering{\includegraphics [width=.40\textwidth] {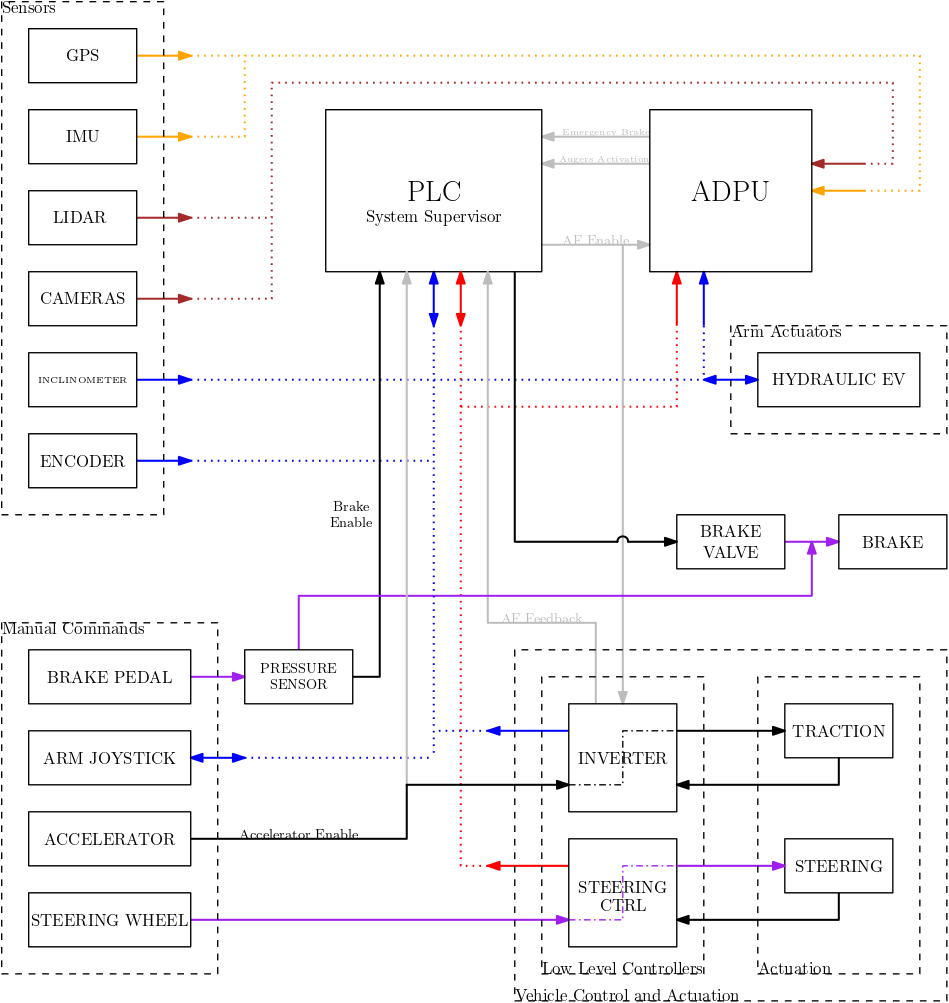}}
\caption{Hardware Architecture of the system in its manually operated configuration, with relevant interconnections.}
\label{fig:HWArchitecture_Manual}
\end{figure}

In MS, the Hardware Architecture (Figure (\ref{fig:HWArchitecture})) undergoes slight modifications, depicted in Figure (\ref{fig:HWArchitecture_Manual}). 
Communication between ADPU and actuators becomes unidirectional, enabling manual override of actuators while maintaining sensor readings and feedback. This setup ensures instantaneous responses to operator inputs prioritizing safety. Actuator control shifts to manual inputs, with hydraulic coupling for the steering wheel and electric coupling for the accelerator pedal. Engagement of manual actuators prompts FSM reset to MS, ensuring coexistence of multiple input solutions without interference.
\section{Preliminary Results}
\label{sec:Results}
\begin{figure}
\centering
\begin{subfigure}{\linewidth}
  \centering
  \includegraphics[height=4.5cm]{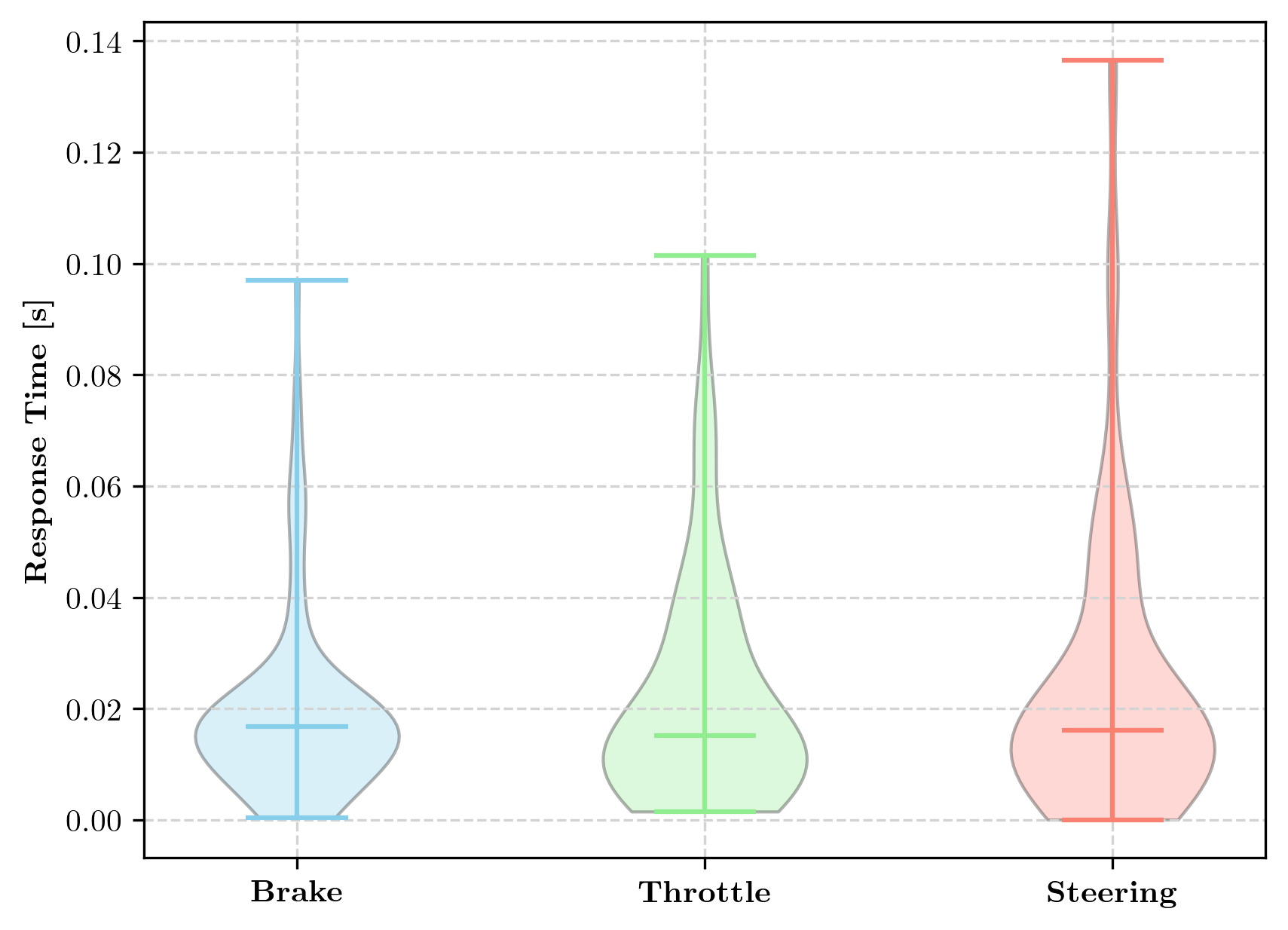}
  \caption{First experimental session.}
  \label{fig:SperimentalData1}
\end{subfigure}
\vspace{.5cm}
\begin{subfigure}{\linewidth}
  \centering
  \includegraphics[height=4.5cm]{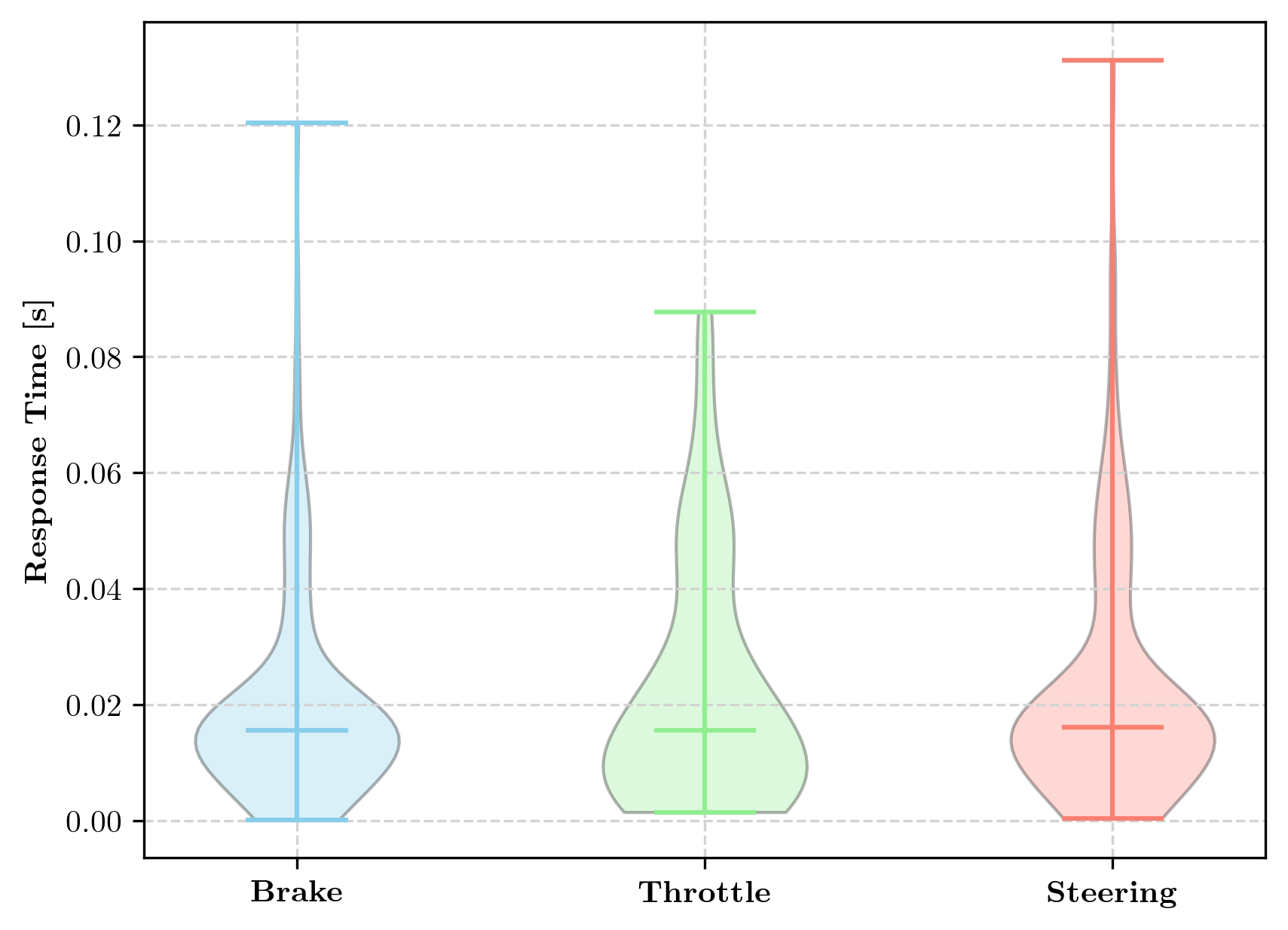}
  \caption{Second experimental session.}
  \label{fig:SperimentalData2}
\end{subfigure}
\vspace{.5cm}
\begin{subfigure}{\linewidth}
  \centering
  \includegraphics[height=4.5cm]{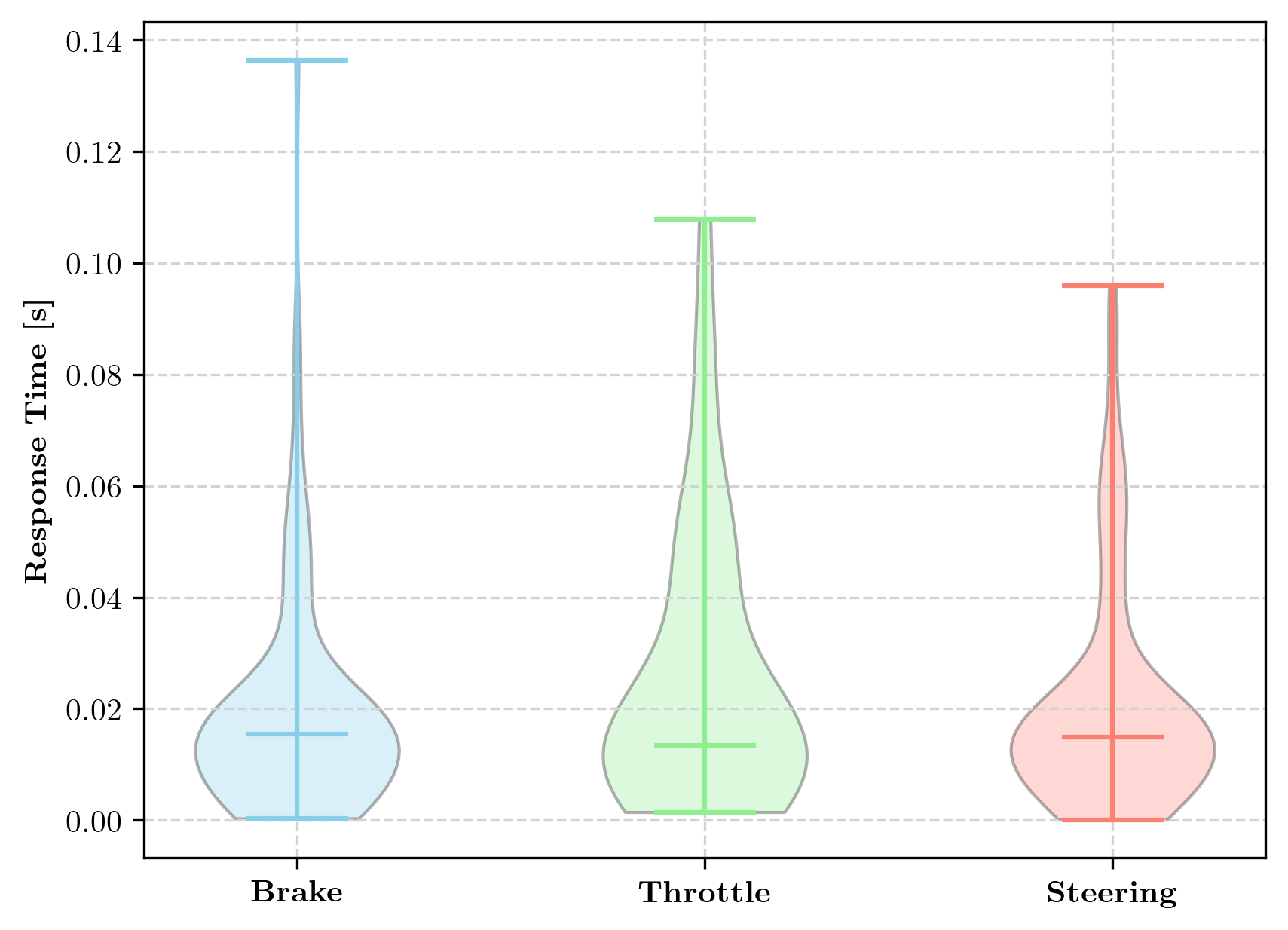}
  \caption{Third experimental session.}
  \label{fig:SperimentalData3}
\end{subfigure}
\caption{Violin plots of system response time during experimental sessions.}
\label{fig:SperimentalData}
\end{figure}
This section presents experimental data from multiple test campaigns with the prototype vehicle, focusing on system response time during manual override maneuvers by operators to assess FSM performance. The objective is to demonstrate the system's controllability by humans in safety-critical scenarios.

Figures \ref{fig:SperimentalData1}, \ref{fig:SperimentalData2}, and \ref{fig:SperimentalData3} illustrate distributions of time intervals between manual command activation during autonomous operations and FSM switching to Manual Control Mode (MS). Data focus on brake pedal, throttle pedal, and steering wheel activation. %Table \ref{tab:CondensedData} summarizes quantities.
%
%
% \begin{table}[ht]
% \centering
% \caption{Descriptive Statistics of Experimental Sessions.}
% \label{tab:CondensedData}
% \begin{tabular}{|c|c|c|c|}
% \hline
%  & Session 1 & Session 2 & Session 3 \\
% \hline
% \multicolumn{4}{|c|}{Brake} \\
% \hline
% Max & $0.0970$ & $0.1204$ & $0.13645$ \\
% Min & $0.0045$ & $0.0015$ & $0.00303$ \\
% Median & $0.01675$ & $0.0155$ & $0.0154$ \\
% \hline
% \multicolumn{4}{|c|}{Throttle} \\
% \hline
% Max & $0.10015$ & $0.08775$ & $0.1079$ \\
% Min & $0.0015$ & $0.00145$ & $0.00145$ \\
% Median & $0.01522$ & $0.01552$ & $0.0134$ \\
% \hline
% \multicolumn{4}{|c|}{Steering} \\
% \hline
% Max & $0.13660$ & $0.1312$ & $0.09595$ \\
% Min & $0.00113$ & $0.0040$ & $0.00157$ \\
% Median & $0.01615$ & $0.0161$ & $0.01485$ \\
% \hline
% \end{tabular}
% \end{table}

These graphs represent distinct experimental sessions with consistent FSM but prototype adjustments in higher-level algorithms (e.g., mapping, localization, planning). Response times for each actuator are uniformly bounded across sessions. Notably, throttle pedal response, despite electrical coupling with the traction motor controller, is comparable to hydraulic commands. Findings suggest response time is linked to system supervisor computational power and communication bus speeds.

Results indicate low-level system behavior remains unaffected by higher-level changes, fulfilling the study's primary goal. The architecture facilitates seamless integration of human and computer inputs without interference.
% \begin{table}[ht]
%     \centering
%     \caption{Descriptive Statistic of the first experimental session.}
%     \label{tab:SperimentalData1}
%     \begin{tabular}{|c|c|c|c|}
%     \hline
%     & Brake & Throttle & Steering \\
%     \hline
%          max & $0.0970$ & $0.10015$ & $0.13660$ \\
%          min & $0.0045$ & $0.0015$ & $0.00113$\\
%          median & $0.01675$ & $0.01522$ & $0.01615$\\
%     \hline
%     \end{tabular}
% \end{table}
%
% \begin{table}[ht]
%     \centering
%     \caption{Descriptive Statistic of the second experimental session.}
%     \label{tab:SperimentalData2}
%     \begin{tabular}{|c|c|c|c|}
%     \hline
%     & Brake & Throttle & Steering \\
%     \hline
%          max & $0.1204$ & $0.08775$ & $0.1312$ \\
%          min & $0.0015$ & $0.00145$ & $0.0040$\\
%          median & $0.0155$ & $0.01552$ & $0.0161$\\
%     \hline
%     \end{tabular}
% \end{table}
%
% \begin{table}[ht]
%     \centering
%     \caption{Descriptive Statistic of the third experimental session.}
%     \label{tab:SperimentalData3}
%     \begin{tabular}{|c|c|c|c|}
%     \hline
%     & Brake & Throttle & Steering \\
%     \hline
%          max & $0.13645$ & $0.1079$ & $0.09595$ \\
%          min & $0.00303$ & $0.00145$ & $0.00157$\\
%          median & $0.0154$ & $0.0134$ & $0.01485$\\
%     \hline
%     \end{tabular}
% \end{table}
%
\section{Conclusion}
\label{sec:Conclusion}
This paper presents a comprehensive exploration of transitioning from traditional human-operated vehicles to fully autonomous solutions tailored for the primary aluminum industry. By amalgamating insights from prior research and integrating state-of-the-art technologies, it addresses critical challenges concerning autonomy, safety, and operational efficiency in industrial settings.

The detailed analysis of system requirements, hardware architecture, software framework and preliminary results significantly contributes to the field of industrial autonomous systems.% By elucidating the complexities of designing, developing, and implementing autonomous features in human-operated vehicles, this research holds implications for enhancing safety and efficiency across various industrial sectors.

Future work could concentrate on refining system architecture, optimizing control algorithms, and enhancing the resilience of autonomous functionalities. These advancements will be pivotal in advancing the adoption of autonomous technologies, fostering innovation, and propelling progress in industrial automation.
\section*{Acknowledgment}
This research is supported by the collaboration of Techmo Car S.p.a. (Padova, Italy) with the University of Padova, Italy.
\bibliography{bibliography.bib} 
\bibliographystyle{ieeetr}
\end{document}